\documentclass[letterpaper, 10 pt, conference]{ieeeconf}
\IEEEoverridecommandlockouts
\usepackage{amsmath,amsfonts,amssymb}
\usepackage{algorithmic}
\usepackage{algorithm}
\usepackage{array}
\usepackage{bbm}
\usepackage[caption=false,font=footnotesize,labelfont=footnotesize,textfont=footnotesize]{subfig}
\usepackage{textcomp}
\usepackage{stfloats}
\usepackage{url}
\usepackage{verbatim}
\usepackage{graphicx}
\usepackage{booktabs}
\usepackage{color}
\usepackage[table]{xcolor}
\usepackage{diagbox}
\usepackage{multirow}
\usepackage{cite}
\usepackage{flushend}
\usepackage{threeparttable}
\makeatletter
\let\NAT@parse\undefined
\makeatother
\usepackage{hyperref}
\usepackage{balance}
\hypersetup{pdfstartview=FitH,
            colorlinks=true,
            linkcolor=red,
            anchorcolor=blue,
            citecolor=green
            }

\newcounter{RNum}

\usepackage{soul}
\soulregister{\cite}7
\soulregister{\citep}7
\soulregister{\citet}7
\soulregister{\ref}7
\soulregister{\pageref}7
\IEEEoverridecommandlockouts                              

\definecolor{table_c}{RGB}{233,238,248}

\title{\LARGE \bf
AnyTSR: Any-Scale Thermal Super-Resolution for UAV
}

\author{Mengyuan Li$^{1}$, Changhong Fu$^{1,2,*}$, Ziyu Lu$^{1}$, Zijie Zhang$^{1}$, Haobo Zuo$^{3}$, Liangliang Yao$^{1}$
\thanks{$^{1}$Mengyuan Li, Changhong Fu, Ziyu Lu, Zijie Zhang, and Liangliang Yao are with the School of Mechanical Engineering, Tongji University, Shanghai 201804, China.
{\tt\small changhongfu@tongji.edu.cn}}
\thanks{$^{2}$Changhong Fu is with the Shanghai Key Laboratory of Wearable Robotics and Human-Machine Interaction, Tongji University, Shanghai 201804, China.}
\thanks{$^{3}$Haobo Zuo is with the Department of Computer Science, University of Hong Kong, Hong Kong 999077, China.}
\thanks{\hspace{0.3cm}*Corresponding author}
}

\begin{document}
\maketitle
\thispagestyle{empty}
\pagestyle{empty}

\begin{abstract}

Thermal imaging can greatly enhance the application of intelligent unmanned aerial vehicles (UAV) in challenging environments. 
However, the inherent low resolution of thermal sensors leads to insufficient details and blurred boundaries.
Super-resolution (SR) offers a promising solution to address this issue, while most existing SR methods are designed for fixed-scale SR. 
They are computationally expensive and inflexible in practical applications.
To address above issues, this work proposes a novel any-scale thermal SR method (AnyTSR) for UAV within a single model.
Specifically, a new image encoder is proposed to explicitly assign specific feature code to enable more accurate and flexible representation. 
Additionally, by effectively embedding coordinate offset information into the local feature ensemble, an innovative any-scale upsampler is proposed to better understand spatial relationships and reduce artifacts.   
Moreover, a novel dataset (UAV-TSR), covering both land and water scenes, is constructed for thermal SR tasks.
Experimental results demonstrate that the proposed method consistently outperforms state-of-the-art methods across all scaling factors as well as generates more accurate and detailed high-resolution images. The code is located at \url{https://github.com/vision4robotics/AnyTSR}.
\end{abstract}

\section{Introduction}
\begin{figure}[htbp]
\centering
\setlength{\abovecaptionskip}{-2pt}
\includegraphics[scale=0.52]{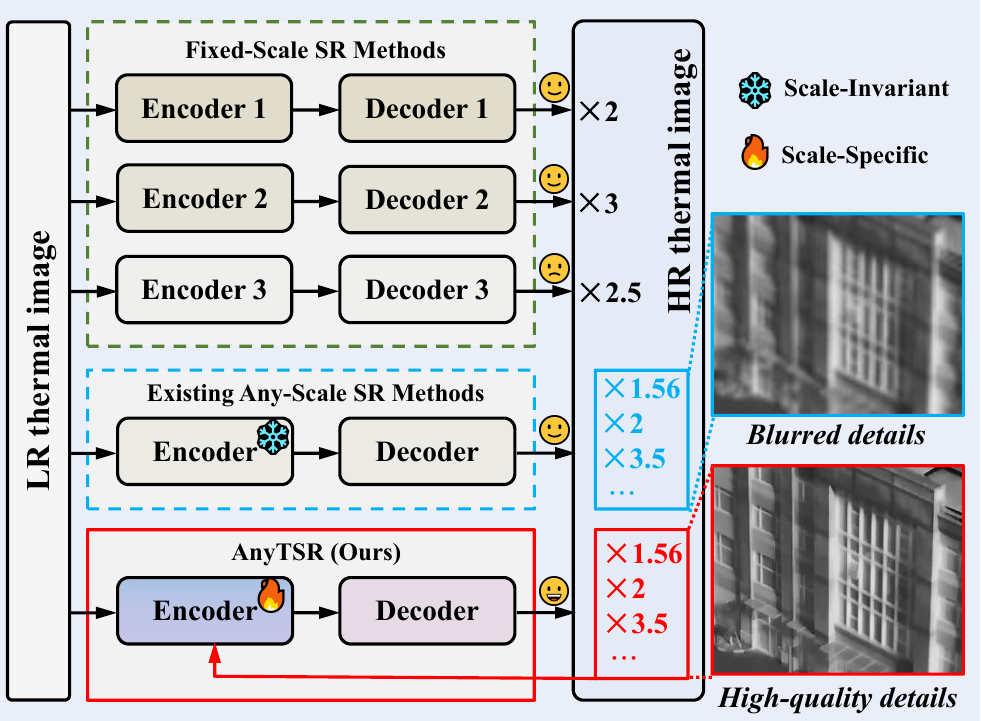}
\caption{Comparison of the proposed AnyTSR with other SR methods. Fixed-scale SR methods require training for specific scales and cannot achieve fractional scaling SR. Existing any-scale SR methods use the same feature representation for all upscaling factors. The proposed method integrates scale-specific information, enabling the model to adaptively understand spatial variations across different scales.}
\label{overview}
\vspace{-15pt}
\end{figure}

Thermal imaging serves as a critical sensing modality for intelligent UAVs \cite{1}, facilitating recognition and tracking in low-light or obscured environments where visible light imaging is ineffective. However, the inherent low resolution (LR) of typical onboard thermal sensors remains a fundamental limitation \cite{4}.
This leads to a reduction in spatial detail, making it difficult to identify small or distant objects, accurately delineate boundaries, and so on. 
High-resolution (HR) thermal sensors onboard UAV platforms are extremely expensive and lack the capability for any-scale zooming.
Super-resolution (SR) technique offers a promising solution to address the resolution limitations, enhancing their spatial quality and enabling more detailed analysis \cite{5}. 
However, as shown in Fig. \ref{overview}, most existing SR methods are designed for fixed-scale SR, where a specific upscaling factor (\textit{e.g.}, ×2, ×4) is predefined during training \cite{7}. These methods require training separate models for each desired scale, which is computationally expensive and inflexible in scenarios that demand adaptive scaling. The limitations of fixed-scale SR methods are 
particularly pronounced in UAV applications, 
where thermal images often need to be processed at varying scales due to changes in altitude and field of view.
\textbf{\textit{Therefore, it is urgent to design a SR framework that can flexibly enhance thermal image resolution at any scales for UAV.}}

Any-scale SR methods have emerged to overcome these challenges \cite{8,26}. These methods are designed to generate high-quality images at any desired resolution without the need for training multiple models. 
Existing any-scale SR methods largely build upon the backbone architectures originally proposed for fixed-scale SR, such as EDSR \cite{9}, RDN \cite{10}, and SwinIR \cite{11}. These models primarily focus on extracting hierarchical features from LR images, and subsequently performing upsampling to achieve the desired resolution. However, 
the SR networks remain entirely unaware of the desired scale before the upsampling operation. The lack of scale-awareness in feature encoding leads to a significant drawback: the same feature representation is used for all upscaling factors, regardless of their specific requirements. Such a design overlooks the fact that different upscaling factors often demand distinct feature representations to effectively capture the unique details and contextual information required for high-quality reconstruction. Consequently, these backbones limit the adaptability and performance of any-scale SR models, especially in scenarios requiring precise and flexible scaling, such as UAV-based thermal imaging.
\textbf{\textit{Thereby, it has been a pressing need to design a scale-aware feature encoding mechanism, ensuring that any-scale SR models generate optimal representations for different scaling requirements.}}

Most existing any-scale SR methods realize the upsampling operation based on implicit image functions \cite{12}, significantly boosting performance compared to traditional interpolation-based methods. These methods predict the RGB value at a query point by aggregating features from its local neighborhood. Typically, they calculate ensemble weights based on the rectangular area between the query point and its nearest feature points. Then, they simply use the weighted features and  coordinate offsets for image reconstruction. Although these method ensures computational simplicity, it introduces several notable limitations. First, the ensemble weights are solely determined by the spatial coordinates of the local features, making them independent of the underlying feature values. Consequently, these methods cannot adaptively adjust the contribution of local features. Second, the separation between feature values and spatial coordinates fail to leverage the rich semantic and structural information embedded in the local features, leading to inaccurate edge reconstruction and the presence of artifacts.
\textbf{\textit{Hence, realizing adaptive feature ensemble to realize offset-aware upsampling, thus improving edge accuracy and reduce artifacts is a pressing need for any-scale SR tasks.}}

In this work, a novel any-scale SR method is proposed to boost the SR performance for thermal images. 
The main contributions are summarized as follows:

\begin{itemize}
\item{A novel any-scale thermal SR network (AnyTSR) for UAV is proposed to realize high-quality image detail reconstruction. Different from most existing methods, our method assigns unique feature codes to different SR scales and thoroughly considers the significance of coordinate offset in SR tasks.}
\item{Scale-specific information is integrated into the state space module, enabling the model to adaptively understand spatial variations across different scales, thereby providing more precise feature representations for the target resolution.}
\item{By explicitly learning feature space weights and embedding coordinate offset information into the local feature ensemble, the proposed method enhances the ability to interpret spatial relationships and accurately reconstruct fine-grained details.}
\item{A novel benchmark (UAV-TSR), covering both land and water scenes, is constructed for thermal SR tasks. Experiments on the proposed dataset demonstrate that the proposed method outperforms existing state-of-the-art methods in both quantitative and qualitative results.}
\end{itemize}

\section{Related Works} \label{Related Work}

\subsection{Single Image Super-Resolution Methods}
Early methods in single image super-resolution relies on patch-based \cite{13} or dictionary-based \cite{14} approaches to model the relationship between high-resolution (HR) and low-resolution (LR) images using external datasets. These traditional techniques are computationally expensive and delivered limited performance. 
C. Dong \textit{et al.} \cite{15} introduces convolutional neural networks (CNNs) into SISR and proposed SRCNN. Later, VDSR \cite{18} improves SR performance by utilizing a very deep convolutional network. RCAN \cite{19} further enhances performance by introducing residual learning and channel attention mechanisms, which accelerates network convergence and enables the model to learn more effective features. Early CNN-based SISR methods typically requires upsampled images as input, leading to excessive computational overhead. To overcome this, FSRCNN \cite{20} utilizes LR images as input and performed upsampling only in the final stages of the network. ESPCN \cite{21} introduces the pixshuffle, which has since been widely adopted. More recently, methods leveraging attention mechanisms have gained prominence. HAN \cite{22} and SAN \cite{23} use spatial attention and second-order attention modules, respectively, to adaptively adjust the output feature maps. SwinIR \cite{11} employs a shifting window technique to capture long-range dependencies across the image. SRFormer \cite{24} introduces Permuted Self-Attention, which enables the advantages of large-window self-attention while effectively reducing computational costs. However, all these methods are tailored for fixed-scale SR and fail to incorporate prior scale information, which restrict their performance in handling any-scale super-resolution tasks.

\subsection{Any-Scale Image Super-Resolution Methods}
Recent advancements have addressed the challenge of any-scale image SR with several methods emerging to improve performance. 
MetaSR \cite{26} introduces the first meta-upscale module for any-scale SR, using EDSR as baselines for feature extraction. Building on this, LIIF \cite{12} utilizes both image coordinates and surrounding 2D deep features as inputs to predict the RGB values at specified coordinates, successfully enhancing any-scale super-resolution. LTE \cite{27}, a dominant-frequency estimator for natural images, enables an implicit function to capture fine details and reconstruct images in a continuous manner. 
CiaoSR \cite{28} employs an implicit attention network that learns ensemble weights for nearby local features and integrates scale-aware attention to exploit additional non-local information. 
SRNO \cite{29} first maps the low-resolution input into a higher-dimensional latent space to capture sufficient basis functions, then iteratively approximates the implicit image function using a kernel integral mechanism, followed by dimensionality reduction to generate the final RGB representation at the target coordinates. However, they use the same feature representation for all upscaling factors. Besides, the separation between feature values and spatial coordinates prevents the model from fully utilizing both visual and positional information for accurate prediction.

\section{Methodology} 
\label{Methodology}
\subsection{Overall Architecture}

\begin{figure*}
\setlength{\abovecaptionskip}{-2pt}
\centering
\includegraphics[width=1\textwidth]{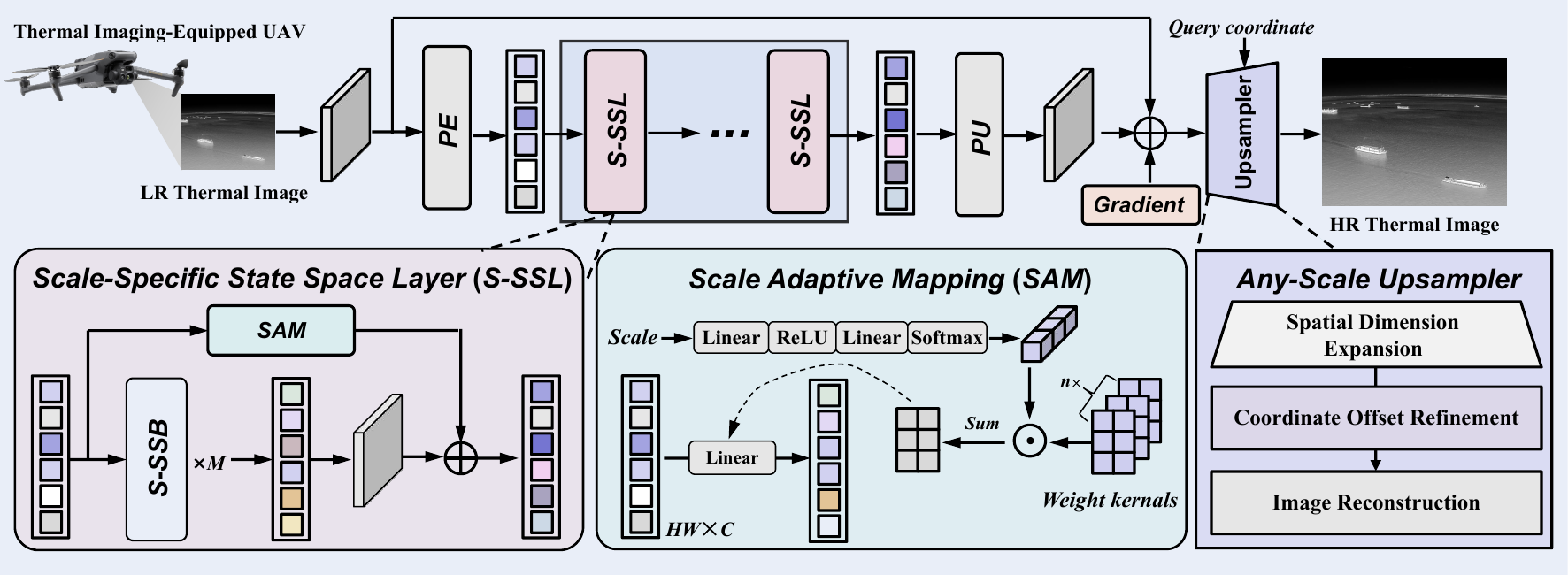}
\caption{The overall architecture of the proposed AnyTSR. The enconder integrates unique scale-specific information to explicitly assign specific feature code to different scale requirements, enabling more accurate and flexible representation across any scales. PE and PU denote patch embedding and patch unembedding. The any-scale upsampler predicts the HR thermal image from the deep latent code.}
\label{fig 1}
\vspace{-15pt}
\end{figure*}

As illustrated in Fig. \ref{fig 1}, the proposed AnyTSR is primarily composed of two parts: the scale-specific LR image encoder $\operatorname{E}_\mathcal{X}$ and the any-scale upsampler $\operatorname{U}$. Specifically, for a given LR thermal image $\textbf{I}^{LR} \in \mathbb{R}^{h \times w \times 1}$, where $h$ and $w$ represent the height and width, respectively. The encoder is used to obtain the deep latent feature code $\textbf{E}^{LR}\in \mathbb{R}^{h \times w \times c}$, where $c$ represent the channel number. This process can be denoted as: 
\begin{equation}
\begin{array}{c}
\textbf{E}^{LR} = \operatorname{E}_\mathcal{X}(\textbf{I}^{LR}) \quad ,\\
\end{array}
\end{equation}
where $\operatorname{E}_\mathcal{X}$ begins by applying a $3 \times 3$ convolution to extract shallow features $\textbf{F}_s \in \mathbb{R}^{h \times w \times c}$. The extracted shallow features $\textbf{F}_s$ are subsequently processed through $N$ Scale-specific State Space Layers, generating deep features $\textbf{F}_n^D \in \mathbb{R}^{h \times w \times c}$ at the $n$-th layer ($n \in \{1, 2, \dots, N\}$). Each layer comprises $M$ scale-specific state space blocks followed by a $3 \times 3$ convolution layer. 

To further enhance the extracted features, an additional convolution layer is employed at the end of the $N$ scale-specific state space layers. Finally, the element-wise summation of the deep features $\textbf{F}_N^D \in \mathbb{R}^{h \times w \times c}$ and the shallow features $\textbf{F}_s$ is computed to form part of the input to the upsampler, denoted as $\textbf{F}_L = \textbf{F}_N^D + \textbf{F}_s$. To preserve boundary and texture details, the first-order and second-order gradients of the input low-resolution image $\textbf{I}^{LR}$ are extracted using the Sobel and Laplacian operators. The gradients are concatenated and subsequently mapped to high-dimensional feature space via a convolution operation. The features are then combined with $\textbf{F}_L$, producing the final deep latent feature code, referred to as $\textbf{E}^{LR}$. $\textbf{E}^{LR}$ serves as the foundation for the subsequent HR image reconstruction.
\begin{align}
\textbf{F}_{\nabla} &= \operatorname{Conv}\left(\operatorname{Cat}\left(\textbf{I}_{\nabla_{1x}}^{LR}, \textbf{I}_{\nabla_{1y}}^{LR}, \textbf{I}_{\nabla_{2}}^{LR}\right)\right) 
\quad ,\\
\textbf{E}^{LR} &= \textbf{F}_{D}^{N} + \textbf{F}_{s} + \textbf{F}_{\nabla} \quad ,
\end{align}
where $\textbf{F}_{\nabla}$ represents the mapped high-dimensional feature of the first-order and second-order gradients. $\textbf{I}_{\nabla_{1x}}^{LR}$ and $\textbf{I}_{\nabla_{1y}}^{LR}$ denote the first-order gradient image along the $x$ and $y$ directions. $\textbf{I}_{\nabla_{2}}^{LR}$ denotes the second-order gradient image. $\operatorname{Cat}(\cdot)$ denotes the concatenation along the channel dimension.

For the given HR query coordinates $\textbf{x}_q \in \textbf{X}^{HR}$, the Any-Scale Upsampler $\operatorname{U}$, leveraging scale prior information $s \in \mathbb{R}^{1}$, predicts the HR thermal image $\textbf{I}^{HR} \in \mathbb{R}^{H \times W \times 1}$ from the deep latent code $\textbf{E}^{LR}$, where $H$ and $W$ represent the HR height and width, respectively. This process consists of three main steps: spatial dimension expansion, coordinate offset refinement, and image reconstruction. Firstly, the spatial dimension of $\textbf{E}^{LR}$ and its associated coordinates $\textbf{x} \in \textbf{X}^{LR}$ are expanded using nearest interpolation to align with the HR query coordinates. Then, the coordinate offsets $\delta \textbf{x}$ are used to predict local ensemble weights ${w}$. Next, along with the weighted HR deep latent code, ${w}$ is utilized to conduct feature refinement. Finally, the reconstructed HR image $\textbf{I}^{HR}$ is generated by applying a neural operator that integrates 3 key elements: the feature offset, the HR deep latent code, and the SR scale. The implementation details of the proposed upsampler are discussed in Sec. \ref{Any-scale Upsampler}.

\noindent\textit{\textbf{Remark 1:}}
The Scale-specific LR Image Encoder is utilized to generate more accurate and flexible representation across any scales. The Any-Scale Upsampler predicts the HR thermal image across any upsampling factors.
\vspace{-5pt}
\subsection{Scale-Specific State Space}
In the context of any-scale SR for thermal images, a key challenge lies in effectively adapting the model to handle varying scale factors. Traditional SR methods struggle with this due to their reliance on fixed feature extraction models. These limitations hinder the ability to adapt to different scale requirements, especially in applications where thermal images need to be processed at varying resolutions due to changes in the position or mission requirements of UAVs. Besides, while Transformer-based architectures \cite{30} excel in many tasks, they face significant challenges with high computational complexity and memory requirements. Mamba \cite{31}, a novel architecture built upon the state-space model, addresses these limitations by improving efficiency in capturing long-term dependencies and enabling the handling of longer sequences.

\begin{figure}[htbp]
\centering
\setlength{\abovecaptionskip}{-2pt}
\includegraphics[scale=0.55]{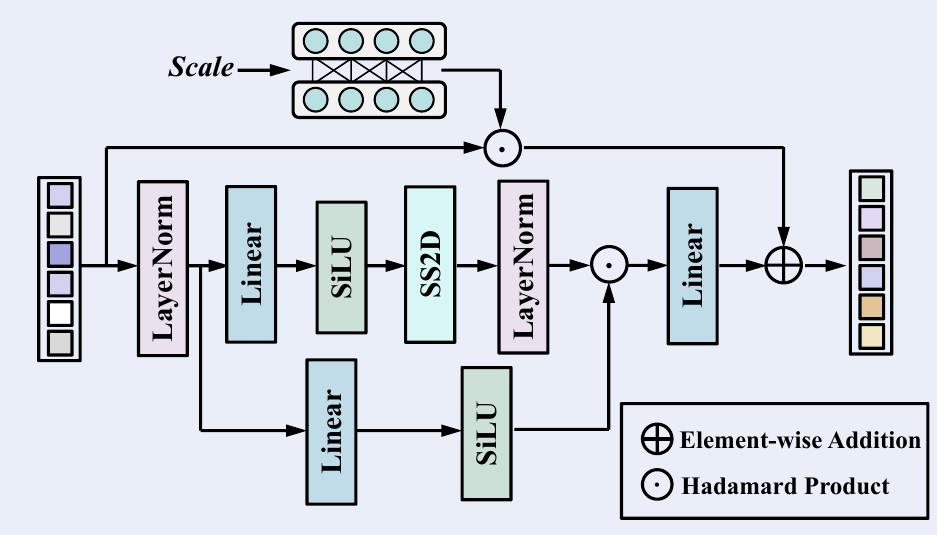}
\caption{The structure of the scale-specific state space block (S-SSB).}
\label{fig 2}
\vspace{-15pt}
\end{figure}
To adapt this approach for any-scale SR tasks, scale-specific state space blocks (S-SSB) is proposed, as shown in Fig. \ref{fig 2}, and scale-specific state space layer (S-SSL) builds upon S-SSB, which integrate SR scale information into the feature transformation process. This allows the network to dynamically adapt to varying SR scales. Let the input feature be $\textbf{F}_{\text{in}} \in \mathbb{R}^{h \times w \times c}$. $\textbf{F}_{\text{in}}$ is processed through Layer Normalization to obtain $\textbf{F}'_{\text{in}}\in \mathbb{R}^{h \times w \times c}$. The normalized feature $\textbf{F}'_{\text{in}}$ is then fed into two separate branches. In the first branch, $\textbf{F}'_{\text{in}}$ is transformed through a linear projection followed by a SiLU activation function:
\begin{equation}
\textbf{F}_{\text{in}}^{1'} = \sigma\left(\operatorname{Linear}(\textbf{F}'_{\text{in}})\right)
\quad ,
\end{equation}
where $\sigma$ denotes the SiLU activation function. In the second branch, $\textbf{F}'_{\text{in}}$ undergoes a depthwise convolution, linear projection, and SiLU activation. It is then processed through a 2D-selective scan module (SS2D) \cite{31}, which dynamically extracts both local and global dependencies. This operation refines the feature representation, producing:
\begin{equation}
\textbf{F}_{\text{in}}^{2'} = \operatorname{LN}\left(\operatorname{SS2D}\left(\sigma\left(\operatorname{DWConv}\left(\operatorname{Linear}(\textbf{F}'_{\text{in}})\right)\right)\right)\right)
\quad ,\end{equation}
where $\operatorname{LN}(\cdot)$ represents Layer Normalization. $\operatorname{DWConv}(\cdot)$ denotes depthwise convolution. To accommodate varying SR scales and make the module well-suited for any-scale SR, the SR scale factor ${s} \in \mathbb{R}^1$ is first transformed into a learnable embedding via a fully connected layer. The scale embedding is then broadcast across spatial dimensions and fused with the input feature through element-wise multiplication, generating the scale-specific feature representation:
\begin{equation}
\textbf{F}_{\text{scale}} = \textbf{F}_{\text{in}} \cdot \operatorname{MLP}({s})
\quad ,
\vspace{-5pt}
\end{equation}
where $\operatorname{MLP}(\cdot)$ represents the fully connected layer. Finally, the output feature of the Scale-specific State Space Block $\textbf{F}_{\text{out}}$ is computed as:
\begin{equation}
\textbf{F}_{\text{out}} = \operatorname{Linear}(\textbf{F}_{\text{in}}^{1'} \cdot \textbf{F}_{\text{in}}^{2'}) + \textbf{F}_{\text{scale}}
\quad .
\vspace{-5pt}
\end{equation}

\noindent\textit{\textbf{Remark 2:}}
By modulating features based on the scale, the block learns to differentiate and emphasize scale-dependent features, enhancing its ability to reconstruct high-quality images at various resolutions.

\subsection{Scale-Specific State Space Layer}
The proposed scale-specific state space layer (S-SSL) builds upon S-SSB, introducing a residual branch called scale adaptive mapping (SAM) to further enhance feature representation. This branch enhances scale-specific feature representation by dynamically learning scale-aware kernel weights. For the given scale factor ${s}$, the SAM branch consists of two key components: kernel weight generation and kernel-based feature transformation. ${s}$ is first utilized to generate $n$ adaptive kernel weights:
\begin{equation}
\textbf{W}_{\text{kernel}} = \operatorname{Softmax}\left(\operatorname{Linear}\left(\operatorname{ReLU}\left(\operatorname{Linear}({s})\right)\right)\right)
\quad .\end{equation}

 A bank of $n$ linear weight kernels with spatial size $C \times C$, denoted as $\textbf{W} \in \mathbb{R}^{n \times C \times C}$, is then combined with the generated kernel $\textbf{W}_{\text{kernel}}$ using a weighted sum across all $n$ kernels. The combined weights are used as the parameters of a linear mapping applied to $\textbf{F}_{\text{in}}$. The final output of the S-SSL is given by combining the outputs of the stacked $M$ S-SSBs and the Scale Adaptive Mapping branch:
\begin{equation}
\textbf{F} = \operatorname{Linear}\left[\sum_{i=1}^{n} \textbf{W}_{\text{kernel}}^{i} \cdot \textbf{W}^{i} ; \textbf{F}_{\text{in}}\right] + \textbf{F}_{\text{out}}^{M}
\quad .\end{equation}

\noindent\textit{\textbf{Remark 3:}}
By incorporating SAM, the S-SSL enhances the ability of scale-specific modeling, ensuring that the extracted features are both content-aware and scale-aware. This design is particularly beneficial for any-scale image SR, as it dynamically adapts to varying scales while maintaining high spatial fidelity.

\begin{figure*}
\centering
\setlength{\abovecaptionskip}{-2pt}
\includegraphics[width=1\textwidth]{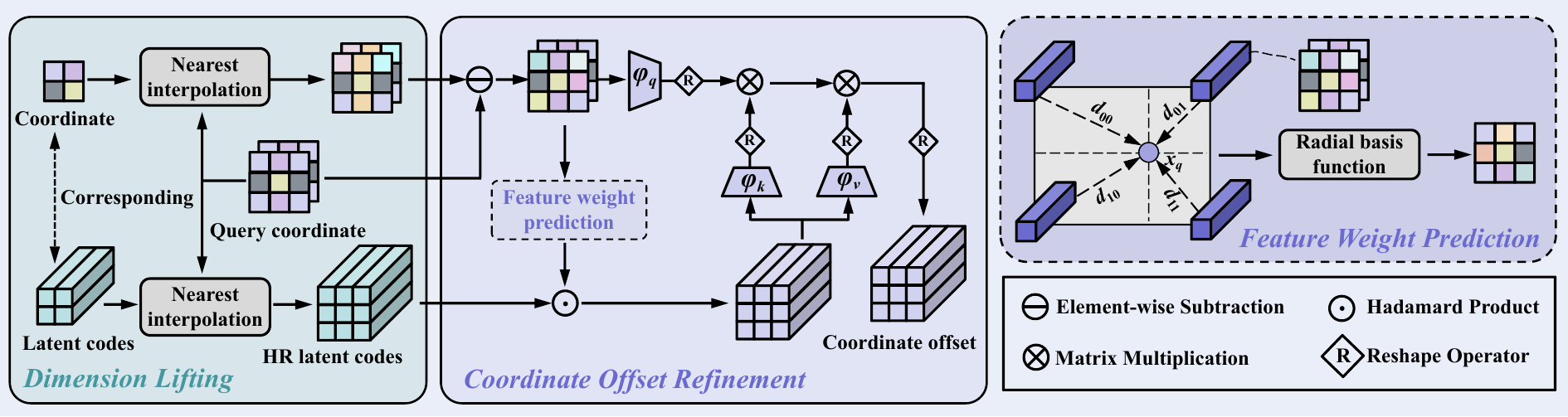}
\caption{The details of the spatial dimension expansion and offset refinement module. The spatial dimension of the deep latent code and its associated coordinates are expanded to align with the HR query coordinates. Coordinate offsets are used to compute weights through a radial basis function (RBF).}
\label{fig 3}
\vspace{-15pt}
\end{figure*}

\vspace{-5pt}
\subsection{Any-Scale Upsampler}
\label{Any-scale Upsampler}
Traditional upsampling methods typically utilize the areas between the target point and its surrounding four points to weight latent codes. While these methods are computationally simple, they struggle to capture the complex relationships between the offsets and the latent codes, resulting in limited detail representation. To address this issue, a novel any-scale upsampler is designed. For the given HR query coordinates $\textbf{x}_q$, the proposed upsampler leverages feature offsets and scale prior information to predict the HR thermal image from latent feature codes. This process consists of three main steps: spatial dimension expansion, coordinate offset refinement, and image reconstruction.
As shown in Fig. \ref{fig 3}, nearest interpolation is used to lift the code dimension according to the latent codes of the four known points surrounding the query coordinates (top left, top right, bottom left, bottom right). The process for each position is the same. Take the top left point as an example, the spatial dimension of the deep latent code $\textbf{E}^{LR}$ and its associated coordinates $\textbf{x}$ are expanded to align with the HR query coordinates $\textbf{x}_q$. Next, let $\textbf{x}'$ be the lifted HR query coordinates, and the coordinate offset values $\delta \textbf{x}$ are used to compute weights through a radial basis function (RBF) \cite{33}. 
\begin{equation}
\delta \textbf{x} = |\textbf{x}_q-\textbf{x}'|
\quad .\end{equation}

Considering that Gaussian Function is popular for its smoothness and effectiveness in interpolation tasks, Gaussian Function is chosen as the RBF. Therefore, the weight of the lifted latent code can be formulated as:
\begin{equation}
{w}_{ij} = \varphi({d}_{ij}) = \exp\left(-\frac{{d}_{ij}^2}{2 \sigma^2}\right)
\quad ,\end{equation}
where $i$ and $j$ represent the index of the spatial position, and ${d}$ represents the Euclidean distance. $\sigma$ is modeled as a learnable parameter. Symbol $\exp(\cdot)$ is the exponential function with base ${e}$. The advantage of using RBF is that, for any given 
${d}$, the predicted ${w}$ always lies between 0 and 1, which shares similarities with area-based methods but brings better generalization. Then, the corresponding latent code is weighted as:
\begin{equation}
\textbf{E}_{ij} = {w}_{ij} \cdot \textbf{E}^{HR}_{ij}
\quad ,\end{equation}
where $\textbf{E}^{HR}$ denotes the lifted HR latent code. By employing RBF to dynamically compute the weighting coefficients based on the distance between the offset and the neighboring points, this method enables more flexible and precise weight assignment compared to traditional area-based weighting. This allows the model to adapt more effectively to local feature variations.
Based on the weighted feature map, an offset refinement module is designed to model the deep interactions between $\delta \textbf{x}$ and $\textbf{E}$. First, two independent convolutional operations are used to generate key vectors $\textbf{K} \in \mathbb{R}^{C \times {HW}}$ and value vectors $\textbf{V} \in \mathbb{R}^{C \times {HW}}$. Then, the query vector $\textbf{Q} \in \mathbb{R}^{C \times {HW}}$, derived from $\delta \textbf{x}$, is combined with $\textbf{K}$ to compute similarity scores. These scores are normalized to produce an attention matrix. Finally, the attention matrix is applied to the value vector $\textbf{V}$, resulting in the learned feature offset $\textbf{E}_{\text{offset}}$.
\begin{align}
\textbf{Q}, \textbf{K}, \textbf{V}&= \varphi_q(\delta \textbf{x}), \varphi_k(\textbf{E}), \varphi_v(\textbf{E})
\quad ,\\
\textbf{E}_{\text{offset}} &= \operatorname{Softmax}(\textbf{Q} \textbf{K}^\text{T}) \textbf{V}
\quad ,\end{align}
where $\varphi_q(\cdot)$, $\varphi_q(\cdot)$, and  $\varphi_v(\cdot)$ represent $1 \times 1$ convolution layers followed by a reshape operation. The feature offset effectively enhances the ability to dynamically capture the relationship between the offset and the feature map. Through the synergistic operation of this module, the proposed module significantly improves the utilization of offset information, enhances the detailed representation of low-resolution feature maps, and addresses the limitations of traditional methods in modeling complex feature relationships, providing superior generalization for SR tasks. Finally, neural operator proposed in \cite{29} is used to realize image reconstruction.

\begin{equation}
\textbf{I}^{HR} = \operatorname{NEO} \Big( {s}, \big\{ \textbf{E}_l, \textbf{E}^l_\text{offset} \big\}_{l=1}^{4} \Big)
\quad ,\end{equation}
where $\operatorname{NEO}$ denotes neural operator, and $\small\{ \cdot\small\}_{l=1}^{4}$ denotes four nearest grid points.

\noindent\textit{\textbf{Remark 4:}}
By explicitly learning feature space weights and embedding coordinate offset information into the local feature ensemble, the proposed method enhances the ability to interpret spatial relationships and accurately reconstruct fine-grained details.
\begin{figure}[t]
\centering
\setlength{\abovecaptionskip}{-2pt}
\includegraphics[scale=0.7]{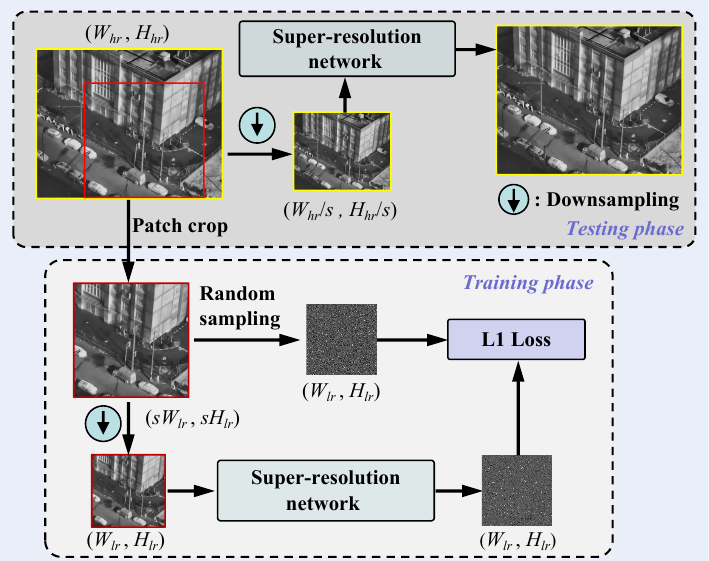}
\caption{The details of the training and testing phase for any-scale super-resolution methods.}
\label{fig 4}
\vspace{-15pt}
\end{figure}

\begin{table*}[hb]
\setlength{\abovecaptionskip}{0.cm}
\caption{Quantitative comparison on the proposed two test sets (PSNR (dB)/LPIPS). Higher PSNR and lower LPIPS indicate better performance. The best performance are bolded. All methods except Bicubic are trained with continuous random scales uniformly sampled in $\times$1 - $\times$4. Test set 1 indicates that the collected images are from land scenes, while Test set 2 indicates that the collected images are from water surface scenes. * indicates that the scale is out of the training distribution.}
\label{table 1}
\centering
\setlength{\abovecaptionskip}{-2pt}
\colorbox{table_c}{
\resizebox{1\textwidth}{!}{  
{\Large
\begin{tabular}{l | c c c c c | c c c c c}
\hline
\multirow{2}{*}{Scales} & 
\multicolumn{5}{c |}{Test set 1} & \multicolumn{5}{c}{Test set 2} \\
\cline{2-11}
 & Bicubic & LIIF \cite{12} & LTE \cite{27} & SRNO \cite{29} & \textbf{Ours} & Bicubic & LIIF \cite{12} & LTE \cite{27} & SRNO \cite{29} & \textbf{Ours} \\
\midrule
$\times$1.45 & 37.49 / 0.0605 & 44.15 / 0.0038 & 44.70 / 0.0030 & 44.80 / 0.0028 & \textbf{45.10 / 0.0027} & 35.77 / 0.0731 & 37.73 / 0.0200 & 37.68 / 0.0193  & 37.94 / 0.0183 & \textbf{38.03 / 0.0173}\\
$\times$1.55 & 36.81 / 0.0700 & 43.26 / 0.0053 & 43.71 / 0.0044 & 43.80 / 0.0041 & \textbf{44.07 / 0.0040} & 35.03 / 0.0828 & 36.87 / 0.0290 & 36.82 / 0.0271 & 37.11 / 0.0264 & \textbf{37.25 / 0.0253}\\
$\times$1.7 & 35.98 / 0.0837 & 42.10 / 0.0080 & 42.44 / 0.0069 & 42.54 / 0.0065 & \textbf{42.78 / 0.0063} & 34.12 / 0.0988 & 35.84 / 0.0449 & 35.81 / 0.0416 & 36.11 / 0.0409 & \textbf{36.26 / 0.0393}\\
$\times$1.8 & 35.50 / 0.0933 & 41.39 / 0.0101 & 41.69 / 0.0089 & 41.79 / 0.0084 & \textbf{42.01 / 0.0082} & 33.59 / 0.1107 & 35.25 / 0.0556 & 35.25 / 0.0517 & 35.57 / 0.0512 & \textbf{35.73 / 0.0493}\\
$\times$2 & 34.61 / 0.1168 & 40.13 / 0.0161 & 40.37 / 0.0144 & 40.49 / 0.0136 & \textbf{40.69 / 0.0129} & 32.66 / 0.1355 & 34.30 / 0.0729 & 34.30 / 0.0681 & 34.59 / 0.0687 & \textbf{34.77 / 0.0663}\\

$\times$2.2 & 33.93 / 0.1433 & 39.04 / 0.0215 & 39.25 / 0.0196 & 39.36 / 0.0190 & \textbf{39.55 / 0.0184} & 31.99 / 0.1625 & 33.51 / 0.0905 & 33.55 / 0.0861 & 33.85 / 0.0866 & \textbf{34.04 / 0.0837}\\
$\times$2.3 & 33.61 / 0.1577 & 38.53 / 0.0251 & 38.73 / 0.0230 & 38.84 / 0.0224 & \textbf{39.03 / 0.0216} & 31.68 / 0.1744 & 33.19 / 0.0978 & 33.25 / 0.0934 & 33.52 / 0.0941 & \textbf{33.73 / 0.0911}\\
$\times$2.45 & 33.16 / 0.1793 & 37.84 / 0.0310 & 38.03 / 0.0286 & 38.15 / 0.0279 & \textbf{38.32 / 0.0269} & 31.24 / 0.1901 & 32.74 / 0.1082 & 32.81 / 0.1041 & 33.06 / 0.1046 & \textbf{33.27 / 0.1018}\\
$\times$2.65 & 32.63 / 0.2056 & 37.01 / 0.0397 & 37.20 / 0.0372 & 37.32 / 0.0362 & \textbf{37.47 / 0.0350} & 30.78 / 0.2072 & 32.28 / 0.1214 & 32.35 / 0.1177 & 32.58 / 0.1182 & \textbf{32.81 / 0.1157}\\
$\times$2.8 & 32.25 / 0.2235 & 36.43 / 0.0468 & 36.61 / 0.0441 & 36.74 / 0.0430 & \textbf{36.89 / 0.0417} & 30.45 / 0.2188 & 31.92 / 0.1306 & 32.00 / 0.1275 & 32.22 / 0.1274 & \textbf{32.44 / 0.1250}\\
$\times$3 & 31.81 / 0.2460 & 35.73 / 0.0551 & 35.92 / 0.0531 & 36.03 / 0.0527 & \textbf{36.17 / 0.0516} & 30.12 / 0.2330 & 31.56 / 0.1413 & 31.63 / 0.1416 & 31.87 / 0.1393 & \textbf{32.10 / 0.1370}\\

$\times$3.1 & 31.58 / 0.2557 & 35.42 / 0.0620 & 35.59 / 0.0595 & 35.72 / 0.0579 & \textbf{35.85 / 0.0565} & 29.94 / 0.2392 & 31.41 / 0.1482 & 31.49 / 0.1459 & 31.69 / 0.1455 & \textbf{31.90 / 0.1434}\\
$\times$3.3 & 31.18 / 0.2761 & 34.81 / 0.0727 & 34.99 / 0.0704 & 35.12 / 0.0685 & \textbf{35.24 / 0.0669} & 29.65 / 0.2524 & 31.07 / 0.1595 & 31.16 / 0.1568 & 31.36 / 0.1567 & \textbf{31.56 / 0.1545}\\
$\times$3.55 & 30.73 / 0.3005 & 34.13 / 0.0862 & 34.30 / 0.0842 & 34.43 / 0.0820 & \textbf{34.55 / 0.0800} & 29.32 / 0.2682 & 30.71 / 0.1727 & 30.80 / 0.1698 & 30.98 / 0.1698 & \textbf{31.19 / 0.1676}\\
$\times$3.7 & 30.48 / 0.3152 & 33.75 / 0.0945 & 33.92 / 0.0927 & 34.05 / 0.0903 & \textbf{34.16 / 0.0882} & 29.14 / 0.2784 & 30.53 / 0.1809 & 30.62 / 0.1778 & 30.79 / 0.1779 & \textbf{31.00 / 0.1758}\\
$\times$3.9 & 30.16 / 0.3348 & 33.28 / 0.1063 & 33.44 / 0.1049 & 33.56 / 0.1024 & \textbf{33.69 / 0.0998} & 28.95 / 0.2895 & 30.30 / 0.1901 & 30.39 / 0.1870 & 30.56 / 0.1870 & \textbf{30.76 / 0.1849}\\
$\times$4 & 30.02 / 0.3467 & 33.06 / 0.1126 & 33.22 / 0.1108 & 33.34 / 0.1083 & \textbf{33.47 / 0.1048} & 28.86 / 0.2971 & 30.19 / 0.1955 & 30.28 / 0.1920 & 30.45 / 0.1918 & \textbf{30.65 / 0.1889}\\

$\times$4.25$^*$ & 29.65 / 0.3653 & 32.53 / 0.1266 & 32.68 / 0.1258 & 32.79 / 0.1233 & \textbf{32.91 / 0.1193} & 28.62 / 0.3087 & 29.91 / 0.2070 & 30.00 / 0.2038 & 30.15 / 0.2037 & \textbf{30.35 / 0.2010}\\
$\times$4.4$^*$ & 29.47 / 0.3774 & 32.25 / 0.1349 & 32.40 / 0.1345 & 32.50 / 0.1320 & \textbf{32.62 / 0.1274} & 28.50 / 0.3148 & 29.76 / 0.2143 & 29.86 / 0.2112 & 30.00 / 0.2107 & \textbf{30.19 / 0.2080}\\
$\times$4.5$^*$ & 29.33 / 0.3837 & 32.05 / 0.1402 & 32.19 / 0.1401 & 32.29 / 0.1376 & \textbf{32.42 / 0.1324} & 28.42 / 0.3151 & 29.67 / 0.2157 & 29.77 / 0.2124 & 29.90 / 0.2119 & \textbf{30.09 / 0.2087}\\
$\times$4.6$^*$ & 29.21 / 0.3932 & 31.89 / 0.1466 & 32.02 / 0.1466 & 32.12 / 0.1443 & \textbf{32.24 / 0.1388} & 28.36 / 0.3242 & 29.59 / 0.2239 & 29.69 / 0.2208 & 29.82 / 0.2201 & \textbf{30.00 / 0.2168}\\
$\times$4.85$^*$ & 28.92 / 0.4081 & 31.46 / 0.1596  & 31.59 / 0.1602 & 31.68 / 0.1579 & \textbf{31.81 / 0.1513} & 28.17 / 0.3291 & 29.37 / 0.2317 & 29.46 / 0.2287 & 29.59 / 0.2277 & \textbf{29.76 / 0.2237}\\
$\times$5$^*$ & 28.74 / 0.4236 & 31.21 / 0.1684 & 31.34 / 0.1699 & 31.42 / 0.1675 & \textbf{31.54 / 0.1603} & 28.09 / 0.3400 & 29.27 / 0.2416 & 29.35 / 0.2392 & 29.47 / 0.2377 & \textbf{29.65 / 0.2329}\\

$\times$5.2$^*$ & 28.53 / 0.4365 & 30.90 / 0.1813 & 31.03 / 0.1827 & 31.11 / 0.1805 & \textbf{31.23 / 0.1720} & 27.98 / 0.3457 & 29.13 / 0.2502 & 29.21 / 0.2473 & 29.33 / 0.2459 & \textbf{29.50 / 0.2406}\\
$\times$5.3$^*$ & 28.42 / 0.4419 & 30.76 / 0.1867 & 30.88 / 0.1885 & 30.96 / 0.1862 & \textbf{31.08 / 0.1773} & 27.90 / 0.3490 & 29.03 / 0.2555 & 29.11 / 0.2524 & 29.24 / 0.2510 & \textbf{29.40 / 0.2452}\\
$\times$5.5$^*$ & 28.23 / 0.4550 & 30.49 / 0.1981 & 30.61 / 0.2004 & 30.69 / 0.1980 & \textbf{30.80 / 0.1876} & 27.82 / 0.3547 & 28.94 / 0.2627 & 29.03 / 0.2596 & 29.13 / 0.2584 & \textbf{29.29 / 0.2513}\\
$\times$5.8$^*$ & 27.95 / 0.4719 & 30.11 / 0.2141 & 30.21 / 0.2172 & 30.29 / 0.2147 & \textbf{30.39 / 0.2020} & 27.67 / 0.3619 & 28.74 / 0.2732 & 28.83 / 0.2701 & 28.93 / 0.2688 & \textbf{29.06 / 0.2596}\\
$\times$5.9$^*$ & 27.86 / 0.4698 & 29.95 / 0.2175 & 30.09 / 0.2209 & 30.16 / 0.2186 & \textbf{30.25 / 0.2052} & 27.58 / 0.3554 & 28.65 / 0.2696 & 28.74 / 0.2666 & 28.84 / 0.2654 & \textbf{28.98 / 0.2555}\\
$\times$6$^*$ & 27.77 / 0.4820 & 29.86 / 0.2258 & 29.96 / 0.2291 & 30.03 / 0.2269 & \textbf{30.12 / 0.2115} & 27.55 / 0.3661 & 28.61 / 0.2813 & 28.70 / 0.2782 & 28.80 / 0.2767 & \textbf{28.92 / 0.2660}\\
\bottomrule
\hline
\end{tabular}
}
}
}
\end{table*}

\vspace{-5pt}
\section{Experiments} \label{Experiment}
\subsection{Implementation Details}

Thermal datasets currently available are predominantly collected for street-view scenarios, with limited options specifically designed for UAV applications. To address this gap, we proposed a novel UAV thermal dataset (UAV-TSR) comprising  1100 HR images, all captured in land scenes. The dataset is systematically divided into 800 images for training and 300 images for testing. Additionally, to further evaluate the generalization capability of SR methods, 200 thermal images over water surfaces are collected as the other test set. This dataset serves as a benchmark for evaluating the performance of any-scale SR methods. These images are captured using the UAV equipped with an integrated thermal imaging module based on an uncooled VOx microbolometer sensor. This thermal camera provides a resolution of 640×512 pixels with a pixel size of 12 µm and operates within a spectral range of 8–14 µm. It supports a frame rate of 30 Hz and has a diagonal field of view of 61°.

Any-scale SR methods are designed to handle varying output resolutions within a single model. This flexibility necessitates additional considerations for batch training, where both input and output resolutions must be fixed. To address this, this work adopts the strategy illustrated in Fig. \ref{fig 4} for training and testing phases. During the training phase, the input resolution $({W}_{lr}, {H}_{lr})$ is fixed at ${W}_{lr}={H}_{lr}=48$. A random scale factor $s$ is sampled uniformly from the range $[1, 4]$, resulting in an upscaled resolution of $({s}{W}_{lr}, {s}{H}_{lr})$. A region of this resolution is then randomly cropped from the HR image to serve as the target output. The input is generated by downsampling this cropped region to $({W}_{lr}, {H}_{lr})$ using bicubic interpolation. To maintain a consistent output resolution for loss calculation, we randomly select ${W}_{lr} \times {H}_{lr}$ pixels from the target output as the ground truth. The model predicts only at these specified coordinates, and the loss is computed using the L1 loss function. To maximize the use of the original images, each image for training is repeated 20 times during the training phase. In the testing phase, the original HR image serves as the ground truth. For a given scale factor ${s}$, the model input is generated by downsampling the HR image by a factor ${s}$ using bicubic interpolation, resulting in an input resolution of $({W}_{hr}/{s}, {H}_{hr}/{s})$.

Adam \cite{34} is used as the optimizer with the initial learning rate $4\times10^{-5}$ and the maximum $4\times10^{-4}$ for training. In order to prevent the instability of the initial training of the model, the warm-up strategy is adopted in the first 20 epochs. The batch size is 16 and the number of training epochs is 100. All models are trained with an NVIDIA A100 GPU (40G Memory). Peak Signal-to-Noise Ratio (PSNR) and Learned Perceptual Image Patch Similarity (LPIPS) are used to measure the model performance. For PSNR, a higher value indicates better performance, whereas for LPIPS, a lower value is preferred.

\subsection{Quantitative Result}
In TABLE \ref{table 1}, several scaling factors are selected randomly from $\times$1-$\times$6 to conduct quantitative comparisons on Test set 1 and Test set 2. The proposed method achieves the highest PSNR values at all testing scales, outperforming both traditional interpolation (Bicubic) and state-of-the-art methods (LIIF \cite{12}, LTE \cite{27}, SRNO \cite{29}). Notably, in Test set 1, at a scaling factor of $\times1.45$, the method attains a PSNR of 45.10 dB, surpassing Bicubic by 7.61 dB and exceeding the second-best method (SRNO) by 0.30 dB. As the scaling factor increases, all methods exhibit gradual PSNR degradation due to amplified reconstruction complexity. However, the proposed method maintains significant advantages even at extreme scales, highlighting its robustness in preserving fine-grained details under challenging conditions. More importantly, the proposed method demonstrates exceptional generalization capability, particularly for unseen scaling factors beyond $\times$4, which were not included during training. Remarkably, at $\times$6, the method attains PSNR/LPIPS of 30.12 dB/0.2115, surpassing SRNO by 0.09 dB/0.0154. 
In Test set 2, the method outperforms all competitors, with PSNR improvements of 0.20–0.46 dB and LPIPS reductions of 0.0029-0.0066 over LIIF, LTE, and SRNO at $\times$4 scales. Notably, for scaling factors beyond the training range, the method demonstrates consistent superiority, attaining 28.92 dB/0.2660 at $\times$6, surpassing SRNO by 0.12 dB/0.0107. The result highlights the robustness of the proposed method, which enables adaptive feature modulation without explicit exposure to extreme scales.
\begin{figure*}[t]
\centering
\setlength{\abovecaptionskip}{-2pt}
\includegraphics[width=1\textwidth]{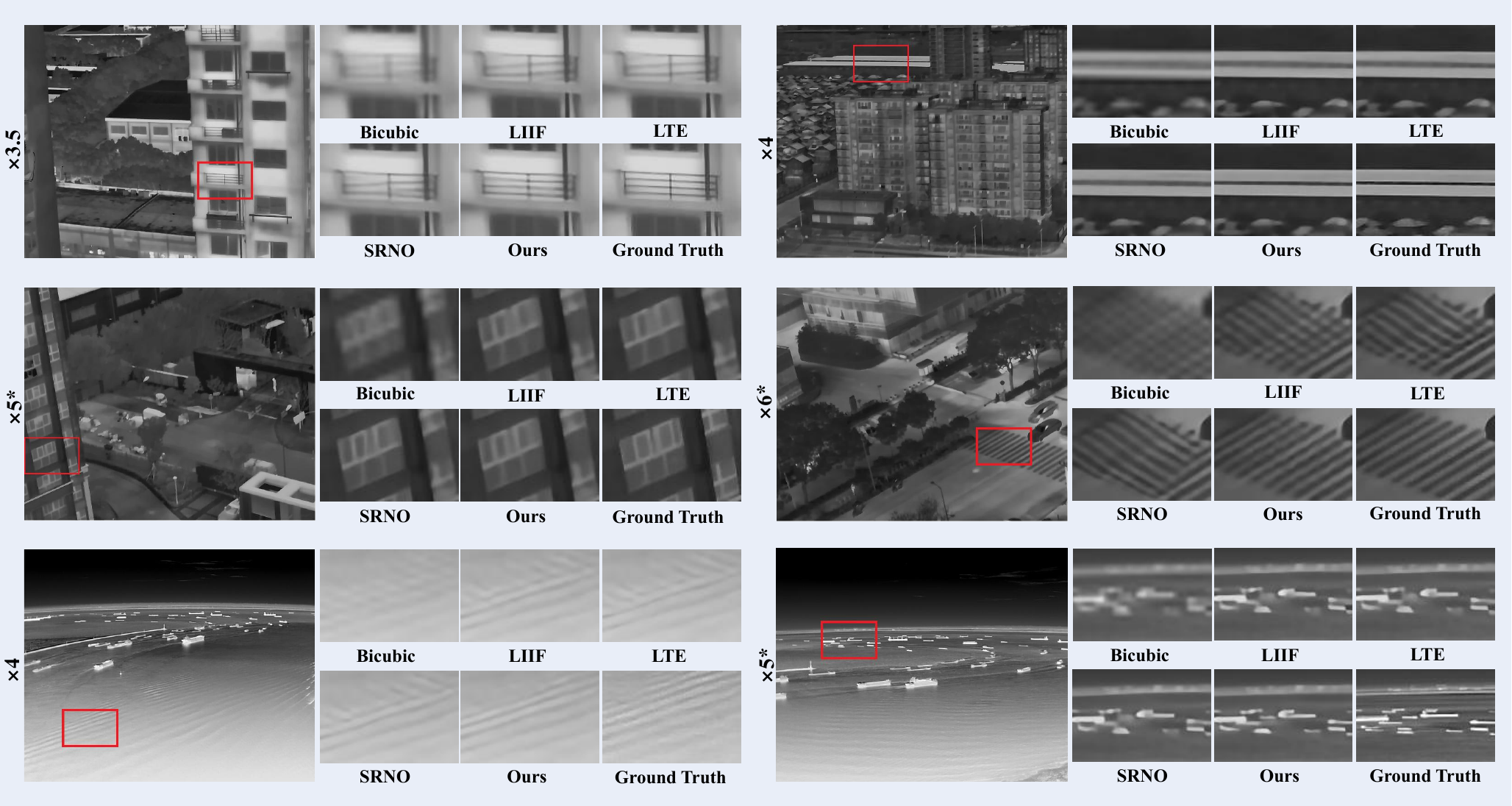}
\caption{The visual results of all any-scale SR methods across multiple scales. The top four images are from Test set 1, and the bottom two images are from Test set 2. * indicates that the scale is out of the training distribution. Ground truth is shown in the bottom right corner. Visualization results demonstrate that the proposed method can generate higher-quality HR images with intricate details.}
\label{fig 5}
\vspace{-17pt}
\end{figure*}
\vspace{-5pt}
\begin{table}[b]
\setlength{\abovecaptionskip}{-2pt}
\caption{Quantitative ablation study of the proposed modules across several scaling factors on Test set 1 (PSNR (DB)).}
\centering
\colorbox{table_c}{
\resizebox{0.45\textwidth}{!}{  
\begin{tabular}{c c c c c c c}
\hline
\textbf{S-SSB} & \textbf{SAM} & \textbf{LLE} & \textbf{ORM} & $\times$2 & $\times$4 & $\times$6 \\
\midrule
& $\checkmark$ & $\checkmark$ & $\checkmark$ & 40.56 & 33.39 & 30.07 \\
$\checkmark$ & & $\checkmark$ & $\checkmark$ & 40.60 & 33.41 & 30.08 \\
$\checkmark$ & $\checkmark$ & & $\checkmark$ & 40.66 & 33.45 & 30.11 \\
$\checkmark$ & $\checkmark$ & $\checkmark$ & $\checkmark$ & \textbf{40.69} & \textbf{33.47} & \textbf{30.12} \\
\hline
\end{tabular}}
\label{table 2}
}
\end{table}
\begin{table}[b]
\setlength{\abovecaptionskip}{-2pt}
\caption{Comparison of PSNR for different synthesis steps.}
\centering
\colorbox{table_c}{
\resizebox{0.45\textwidth}{!}{
\begin{tabular}{l l c c}
\hline
\textbf{Type} & \textbf{Synthesis steps} & \textbf{Test set 1} & \textbf{Test set 2}\\
\hline
Multiple steps & $\rightarrow $2$\rightarrow $4$\rightarrow $6 & 28.01 & 28.03\\
 & $\rightarrow $2$\rightarrow $6 & 28.83 & 28.52\\
\hline
One step & $\rightarrow $6 & \textbf{30.12} & \textbf{28.92}\\
\hline
\end{tabular}}
\label{table 3}
}
\vspace{-10pt}
\end{table}
\subsection{Qualitative Result}
The SR results of several images from the UAV-TSR test set are present in Fig. \ref{fig 5}. Bicubic introduces significant artifacts, especially in regions with fine structures. The buildings and their intricate textures, such as windows and edges, are heavily blurred, leading to a lack of detail in the higher magnification areas. While LTE shows marginal improvements in certain regions through local fourier frequency estimation, artifacts remain dominant overall. SRNO more effectively recovers fine details by utilizing the global integral kernels. 
However, simple modeling of coordinate offsets notably reduces the accuracy of detail reconstruction, particularly at extreme super-resolution scales. This issue is clearly evident in the bottom-right image (×6), where the reconstruction of the zebra crossing is completely misaligned with the ground truth. In contrast, the proposed method ensures precise reconstruction of both object positions and high-frequency edge details. 
The reconstruction of small objects is notably more accurate compared to SRNO, with sharper outlines and better alignment with the ground truth. The fine textures, like the position of the railing and the zebra crossing, are well-preserved, even at extreme scale factors. Additionally, the proposed method demonstrates its clear advantage by preserving fine textures such as the ripples on the water surface. Unlike SRNO, which struggles with these intricate details, the proposed method accurately reconstructs the water texture, maintaining the subtle variations in shading and pattern that are crucial for visual authenticity in natural scenes.
\vspace{-5pt}

\subsection{Ablation Study}
 Ablation experiments are conducted to validate the effectiveness of key components by systematically removing individual modules. All experiments are performed on the UAV-TSR test set. 

TABLE \ref{table 2} demonstrates the progressive impact of each component on PSNR (dB) across scaling factors. S-SSB, SAM, LLE, ORM represent scale-specific state space block, scale adaptive mapping, learnable local ensemble and offset refinement module, respectively. The whole model achieves the highest PSNR values, confirming the synergistic effectiveness of all components. PSNR gradually decreases as modules are removed one by one.


The trained model can synthesize SR images with any given any scale in one step. An alternative is to generate the same scale in multiple steps. Taking a large scale (\textit{e.g.}, $\rightarrow $$\times$6) as an example, SR images can be generated by two steps (\textit{e.g.}, $\rightarrow $$\times$2$\rightarrow $$\times$6), or three steps (\textit{e.g.}, $\rightarrow $$\times$2$\rightarrow $$\times$4$\rightarrow $$\times$6). The results of synthesizing a large scale in one and multiple steps on the two test sets are shown in TABLE \ref{table 3}. It is observed that the way of synthesizing SR images in one step has the best performance. In contrast, the performance becomes worse with more steps because multiple synthesizing steps suffer from error accumulation.

\section{Conclusion} \label{Conclusion}
In this work, a novel any-scale thermal SR network (AnyTSR) for UAV is proposed. By integrating scale-specific information into the state space module and leveraging an innovative any-scale upsampler, our method effectively addresses the limitations of fixed-scale SR methods. This work highlights the potential of scale-aware thermal image SR in UAV applications, offering a promising solution to the challenge of LR thermal images. Moreover, the proposed framework can serve as a teacher model for distillation, enabling the development of lightweight variants suitable for edge deployment. Future research could focus on further optimizing the model for real-time performance and exploring multimodal fusion applications.

\section*{Acknowledgments}
This work was supported by the National Natural Science Foundation of China (62173249, U24B20161) and Natural Science Foundation of Shanghai (20ZR1460100).

\balance
\bibliographystyle{IEEEtran}
\bibliography{reference}

\begin{thebibliography}{10}
\providecommand{\url}[1]{#1}
\csname url@samestyle\endcsname
\providecommand{\newblock}{\relax}
\providecommand{\bibinfo}[2]{#2}
\providecommand{\BIBentrySTDinterwordspacing}{\spaceskip=0pt\relax}
\providecommand{\BIBentryALTinterwordstretchfactor}{4}
\providecommand{\BIBentryALTinterwordspacing}{\spaceskip=\fontdimen2\font plus
\BIBentryALTinterwordstretchfactor\fontdimen3\font minus \fontdimen4\font\relax}
\providecommand{\BIBforeignlanguage}[2]{{%
\expandafter\ifx\csname l@#1\endcsname\relax
\typeout{** WARNING: IEEEtran.bst: No hyphenation pattern has been}%
\typeout{** loaded for the language `#1'. Using the pattern for}%
\typeout{** the default language instead.}%
\else
\language=\csname l@#1\endcsname
\fi
#2}}
\providecommand{\BIBdecl}{\relax}
\BIBdecl

\bibitem{1}
B.~Huang, Z.~Dou, J.~Chen, J.~Li, N.~Shen, Y.~Wang, and T.~Xu, ``{Searching Region-Free and Template-Free Siamese Network for Tracking Drones in TIR Videos},'' \emph{IEEE Transactions on Geoscience and Remote Sensing}, vol.~62, pp. 1--15, 2024.

\bibitem{4}
W.~Yang, H.~Luo, K.-W. Tse, H.~Hu, K.~Liu, B.~Li, and C.-Y. Wen, ``{Autonomous–Targetless Extrinsic Calibration of Thermal, RGB, and LiDAR Sensors},'' \emph{IEEE Transactions on Instrumentation and Measurement}, vol.~73, pp. 1--11, 2024.

\bibitem{5}
R.~Li and X.~Zhao, ``{LSwinSR: UAV Imagery Super-Resolution Based on Linear Swin Transformer},'' \emph{IEEE Transactions on Geoscience and Remote Sensing}, vol.~62, pp. 1--13, 2024.

\bibitem{7}
S.~Wang, T.~Zhou, Y.~Lu, and H.~Di, ``{Contextual Transformation Network for Lightweight Remote-Sensing Image Super-Resolution},'' \emph{IEEE Transactions on Geoscience and Remote Sensing}, vol.~60, pp. 1--13, 2022.

\bibitem{8}
S.~Jiang, N.~Li, M.~Xu, S.~Zhang, and S.~Jia, ``{SQformer: Spectral-Query Transformer for Hyperspectral Image Arbitrary-Scale Super-Resolution},'' \emph{IEEE Transactions on Geoscience and Remote Sensing}, vol.~62, pp. 1--15, 2024.

\bibitem{26}
X.~Hu, H.~Mu, X.~Zhang, Z.~Wang, T.~Tan, and J.~Sun, ``{Meta-SR: A Magnification-Arbitrary Network for Super-Resolution},'' in \emph{Proceedings of the IEEE/CVF Conference on Computer Vision and Pattern Recognition (CVPR)}, 2019, pp. 1575--1584.

\bibitem{9}
B.~Lim, S.~Son, H.~Kim, S.~Nah, and K.~Mu~Lee, ``{Enhanced Deep Residual Networks for Single Image Super-Resolution},'' in \emph{Proceedings of the IEEE Conference on Computer Vision and Pattern Recognition Workshops (CVPRW)}, 2017, pp. 136--144.

\bibitem{10}
Y.~Zhang, Y.~Tian, Y.~Kong, B.~Zhong, and Y.~Fu, ``{Residual Dense Network for Image Super-Resolution},'' in \emph{Proceedings of the IEEE Conference on Computer Vision and Pattern Recognition (CVPR)}, 2018, pp. 2472--2481.

\bibitem{11}
J.~Liang, J.~Cao, G.~Sun, K.~Zhang, L.~Van~Gool, and R.~Timofte, ``{SwinIR: Image Restoration Using Swin Transformer},'' in \emph{Proceedings of the IEEE/CVF International Conference on Computer Vision (ICCV)}, 2021, pp. 1833--1844.

\bibitem{12}
Y.~Chen, S.~Liu, and X.~Wang, ``{Learning Continuous Image Representation with Local Implicit Image Function},'' in \emph{Proceedings of the IEEE/CVF Conference on Computer Vision and Pattern Recognition (CVPR)}, 2021, pp. 8628--8638.

\bibitem{13}
O.~Mac~Aodha, N.~D. Campbell, A.~Nair, and G.~J. Brostow, ``{Patch Based Synthesis for Single Depth Image Super-Resolution},'' in \emph{Proceedings of the European Conference on Computer Vision (ECCV)}, 2012, pp. 71--84.

\bibitem{14}
S.~Yang, Z.~Liu, M.~Wang, F.~Sun, and L.~Jiao, ``{Multitask Dictionary Learning and Sparse Representation Based Single-Image Super-Resolution Reconstruction},'' \emph{Neurocomputing}, vol.~74, no.~17, pp. 3193--3203, 2011.

\bibitem{15}
C.~Dong, C.~C. Loy, K.~He, and X.~Tang, ``{Image Super-Resolution Using Deep Convolutional Networks},'' \emph{IEEE Transactions on Pattern Analysis and Machine Intelligence}, vol.~38, no.~2, pp. 295--307, 2015.

\bibitem{18}
J.~Kim, J.~K. Lee, and K.~M. Lee, ``{Accurate Image Super-Resolution Using Very Deep Convolutional Networks},'' in \emph{Proceedings of the IEEE Conference on Computer Vision and Pattern Recognition (CVPR)}, 2016, pp. 1646--1654.

\bibitem{19}
Y.~Zhang, K.~Li, K.~Li, L.~Wang, B.~Zhong, and Y.~Fu, ``{Image Super-Resolution Using Very Deep Residual Channel Attention Networks},'' in \emph{Proceedings of the European Conference on Computer Vision (ECCV)}, 2018, pp. 286--301.

\bibitem{20}
C.~Dong, C.~C. Loy, and X.~Tang, ``{Accelerating the Super-Resolution Convolutional Neural Network},'' in \emph{Proceedings of the European Conference on Computer Vision (ECCV)}, 2016, pp. 391--407.

\bibitem{21}
W.~Shi, J.~Caballero, F.~Husz{\'a}r, J.~Totz, A.~P. Aitken, R.~Bishop, D.~Rueckert, and Z.~Wang, ``{Real-Time Single Image and Video Super-Resolution Using an Efficient Sub-Pixel Convolutional Neural Network},'' in \emph{Proceedings of the IEEE Conference on Computer Vision and Pattern Recognition (CVPR)}, 2016, pp. 1874--1883.

\bibitem{22}
B.~Niu, W.~Wen, W.~Ren, X.~Zhang, L.~Yang, S.~Wang, K.~Zhang, X.~Cao, and H.~Shen, ``{Single Image Super-Resolution via a Holistic Attention Network},'' in \emph{Proceedings of the European Conference on Computer Vision (ECCV)}, 2020, pp. 191--207.

\bibitem{23}
T.~Dai, J.~Cai, Y.~Zhang, S.-T. Xia, and L.~Zhang, ``{Second-Order Attention Network for Single Image Super-Resolution},'' in \emph{Proceedings of the IEEE/CVF Conference on Computer Vision and Pattern Recognition (CVPR)}, 2019, pp. 11\,065--11\,074.

\bibitem{24}
Y.~Zhou, Z.~Li, C.-L. Guo, S.~Bai, M.-M. Cheng, and Q.~Hou, ``{SRFormer: Permuted self-Attention for Single Image Super-Resolution},'' in \emph{Proceedings of the IEEE/CVF International Conference on Computer Vision (ICCV)}, 2023, pp. 12\,780--12\,791.

\bibitem{27}
J.~Lee and K.~H. Jin, ``{Local Texture Estimator for Implicit Representation Function},'' in \emph{Proceedings of the IEEE/CVF Conference on Computer Vision and Pattern Recognition (CVPR)}, 2022, pp. 1929--1938.

\bibitem{28}
J.~Cao, Q.~Wang, Y.~Xian, Y.~Li, B.~Ni, Z.~Pi, K.~Zhang, Y.~Zhang, R.~Timofte, and L.~Van~Gool, ``{Ciaosr: Continuous Implicit Attention-in-Attention Network for Arbitrary-Scale Image Super-Resolution},'' in \emph{Proceedings of the IEEE/CVF Conference on Computer Vision and Pattern Recognition (CVPR)}, 2023, pp. 1796--1807.

\bibitem{29}
M.~Wei and X.~Zhang, ``{Super-Resolution Neural Operator},'' in \emph{Proceedings of the IEEE/CVF Conference on Computer Vision and Pattern Recognition (CVPR)}, 2023, pp. 18\,247--18\,256.

\bibitem{30}
A.~Waswani, N.~Shazeer, N.~Parmar, J.~Uszkoreit, L.~Jones, A.~Gomez, L.~Kaiser, and I.~Polosukhin, ``{Attention is all you need},'' in \emph{Proceedings of Advances in Neural Information Processing Systems (NIPS)}, 2017, pp. 1--15.

\bibitem{31}
Y.~Liu, Y.~Tian, Y.~Zhao, H.~Yu, L.~Xie, Y.~Wang, Q.~Ye, J.~Jiao, and Y.~Liu, ``{VMamba: Visual State Space Model},'' in \emph{Proceedings of Advances in Neural Information Processing Systems (NIPS)}, vol.~37, 2025, pp. 103\,031--103\,063.

\bibitem{33}
M.~D. Buhmann, ``{Radial Basis Functions},'' \emph{Acta Numerica}, vol.~9, pp. 1--38, 2000.

\bibitem{34}
D.~P. Kingma, ``{Adam: A Method for Stochastic Optimization},'' \emph{arXiv preprint arXiv:1412.6980}, pp. 1--15, 2014.

\end{thebibliography}
\end{document}